%% file: main.tex
\documentclass[10pt,twocolumn,letterpaper]{article}

\usepackage[margin=0.75in,top=1in,bottom=1in,columnsep=0.3125in]{geometry}
\usepackage{times}
\usepackage{xspace}
\usepackage[dvipsnames,table]{xcolor}
\usepackage{graphicx}
\usepackage{amsmath}
\usepackage{amssymb}
\usepackage{booktabs}
\usepackage[numbers,sort&compress]{natbib}
\usepackage[format=plain,labelformat=simple,labelsep=period,font=small,compatibility=false]{caption}
\usepackage[font=footnotesize,skip=3pt,subrefformat=parens]{subcaption}
\usepackage[hyphens]{url}
\usepackage[shortlabels,inline]{enumitem}
\usepackage{multirow}
\usepackage{hyphenat}
\def\ie{i.e.\@\xspace}
\def\eg{e.g.\@\xspace}

\input{preamble}

\definecolor{cvprblue}{rgb}{0.21,0.49,0.74}
\usepackage[breaklinks,colorlinks,allcolors=cvprblue]{hyperref}
\usepackage[capitalize]{cleveref}

\title{AdapTS: Lightweight Teacher-Student Approach for \\ Multi-Class and Continual
Visual Anomaly Detection}

\author{Manuel Barusco\\
University of Padova, Italy\\
{\tt\small manuel.barusco@phd.unipd.it}
\and
Davide Dalle Pezze\\
University of Padova, Italy\\
{\tt\small davide.dallepezze@unipd.it}
\and
Francesco Borsatti\\
University of Padova, Italy\\
{\tt\small francesco.borsatti.1@phd.unipd.it}
\and
Gian Antonio Susto\\
University of Padova, Italy\\
{\tt\small gianantonio.susto@unipd.it}
}

\begin{document}
\maketitle
\input{sec/abstract}

\input{sec/introduction}

\input{sec/related_work}

\input{sec/methodology}

\input{sec/settings}

\input{sec/results}

\input{sec/conclusions}

{
    \small
    \bibliographystyle{IEEEtran}
    \bibliography{main}
}

\end{document}

%% file: preamble.tex
\newcommand{\davide}[1]{\textcolor{black}{#1}}

%% file: sec/abstract.tex
\begin{abstract}
Visual Anomaly Detection (VAD) is crucial for industrial inspection.
However, most methods are designed for single-category models, and Teacher-Student (TS) approaches, despite their efficiency, have not been designed for the Continual scenario, where they need a replay buffer.
To bridge this gap, we propose AdapTS, a unified TS framework for multi-class and continual settings, optimized for resource-constrained deployment.
A single shared frozen backbone acts as teacher and student, and lightweight trainable adapters, selected at inference by a prototype-based task identification (99\% accuracy), store the knowledge of each category without keeping past data.
Training is enhanced via a segmentation-guided objective on synthetic Perlin noise.
On MVTec AD, AdapTS-M improves I-ROC by 11 points over STFPM and needs 4.3$\times$ less memory than Dinomaly, which is 4 points ahead. In the Continual scenario, AdapTS-M reaches the best pixel-level AP with only 0.1\% forgetting and no replay buffer. AdapTS-S needs only 23~MB, and with a MobileNetV2 backbone AdapTS needs 6--7~MB while outperforming the edge-oriented STFPM on all metrics.
AdapTS is thus a viable solution for resource-constrained deployment.
\end{abstract}

%% file: sec/introduction.tex
\section{Introduction}
\label{sec:introduction}

Visual Anomaly Detection (VAD) aims to detect and localize anomalies in industrial images. VAD offers two key advantages: it requires only defect-free images for training, avoiding the cost of collecting and annotating defective samples, and it localizes anomalous regions, providing interpretable outputs for operators. 
Most of the VAD literature considers single-category models, i.e., one model per product, which is costly: training, storage, and maintenance costs grow with every product, and each new product requires a new deployment.
Real inspection lines, however, are better described by two settings: multiple products handled together by a single model (Multi-Class Setting), and new products introduced continuously over time (Continual Setting).
\\
Feature-based approaches, which use pretrained encoders, generally offer the best efficiency-performance trade-off. Among them, teacher-student (TS) approaches such as Dinomaly~\cite{dinomaly} and MambaAD~\cite{he2024mambaad} handle the multi-class setting but do not natively support the continual one, where they need a replay buffer, i.e., a memory of past images, to mitigate forgetting. To the best of our knowledge, no existing TS method addresses both settings in a single framework.
In addition, memory is a key constraint because inspection often runs on devices with limited resources next to the production line and these methods are memory-hungry: Dinomaly and MambaAD require 561~MB and 330~MB, respectively.  
\\
\begin{figure*}[t]
    \centering
    \includegraphics[width=\textwidth]{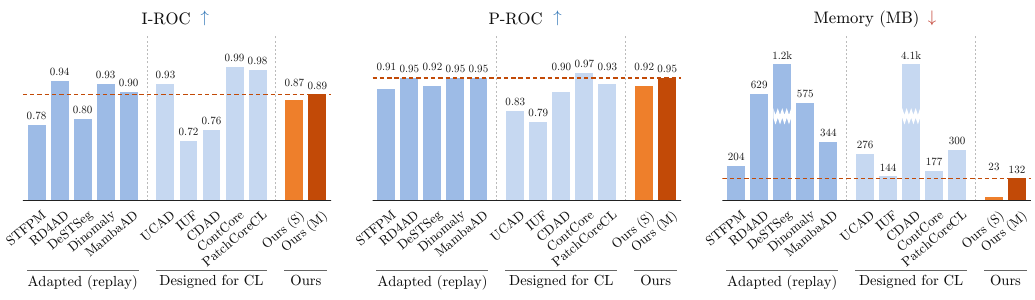}
    \caption{AdapTS (S and M) against the state of the art on MVTec AD in the Continual scenario. Adapted: methods not designed for continual learning, trained with a replay buffer. Designed for CL: methods built for continual learning. The dashed line marks AdapTS-M. Bars exceeding the axis are broken. The ROC axes start at 0.5, the value of a random classifier. The memory of UCAD, IUF, CDAD and ContCore is from the literature.}
    \label{fig:tradeoff}
\end{figure*}
In this work, we propose \textbf{AdapTS}, a TS approach that natively supports both the multi-class and continual settings, with a focus on resource-constrained deployment.
Rather than maintaining two separate networks as in standard TS methods, a single pretrained backbone plays the role of both teacher and student: small trainable adapters are attached to the student pathway, while the backbone weights stay frozen. A segmentation-guided objective on synthetic Perlin-noise anomalies makes the resulting feature discrepancies spatially coherent. 
At inference, a prototype-based task identification mechanism selects the appropriate adapters.
Thanks to its task-specific adapters, AdapTS, unlike Dinomaly and MambaAD, does not need to store past data to work in the continual setting. Moreover, thanks to the shared frozen backbone and lightweight adapters, AdapTS-M requires only 132~MB and AdapTS-S only 23~MB of total memory on MVTec AD.
Our contributions are as follows:
\begin{itemize}
    \item We propose AdapTS, a TS approach for VAD in which a single frozen backbone acts as both teacher and student through lightweight adapters, trained with a segmentation-guided objective whose head is discarded at inference.
    \item Unlike existing TS methods, AdapTS is natively suited for both multi-class and continual learning settings, through task-specific adapters selected via task identification.
    \item We evaluate AdapTS on MVTec AD and VisA. In the Continual scenario on MVTec AD, AdapTS-M reaches the highest pixel-level AP (0.50) with 132~MB, while the other TS methods need between 204 and 1,229~MB and show 7--12\% forgetting against our 0.1\%.
\end{itemize}

%% file: sec/related_work.tex
\section{Related Work}
\label{sec:related_work}

\paragraph{Visual Anomaly Detection.}
Early VAD approaches relied on \textbf{reconstruction-based} methods (autoencoders or GANs), which are computationally expensive and often underperform. The current trend favors \textbf{feature-based} VAD, which leverages pretrained models for higher performance at lower cost. Feature-based methods include memory-bank methods such as PatchCore \cite{patch}, PaDiM \cite{PaDiM}, and CFA \cite{lee2022cfa}, which store representative patch-level embeddings; normalizing flow methods such as FastFlow \cite{yu2021fastflow}, which model the normal feature distribution but are computationally intensive; and TS methods, which use the feature discrepancy between a frozen teacher and a trainable student as the anomaly signal.

\paragraph{Teacher-Student Methods.}
TS methods rely on knowledge distillation: a student trained only on normal samples produces features that diverge from those of a pretrained teacher when anomalous inputs are encountered.
STFPM \cite{wang2021studentteacherfeaturepyramidmatching}, summarized in Sec.~\ref{subsec:STFPM}, is an early and widely used encoder-encoder TS method.
The current state-of-the-art TS methods moved from the encoder-encoder architecture of STFPM to an encoder-decoder one, as in RD4AD, DeSTSeg, Dinomaly, and MambaAD. These approaches combine reconstruction with the TS framework: like TS methods, they use a frozen encoder and detect anomalies by comparing the feature maps of the encoder and of the decoder, rather than the images as in classic reconstruction-based methods. They are computationally cheaper than reconstruction-based models, but heavier than STFPM (Table~\ref{tab:results-multi-eff}).
In RD4AD \cite{deng2022anomaly}, the student receives a compact bottleneck embedding and reconstructs the teacher's multi-scale representations, suppressing anomalous perturbations through reverse distillation. DeSTSeg \cite{zhang2023destseg} trains the student to reconstruct the teacher's clean features from synthetically corrupted images and fuses multi-level discrepancies with a segmentation network supervised by synthetic anomaly masks. Dinomaly \cite{dinomaly} follows RD4AD with a pretrained DINOv2 encoder, unbiased attention layers, loose reconstruction constraints, and a noisy bottleneck that prevents trivial identity reconstruction.
MambaAD \cite{he2024mambaad} instead replaces the CNN decoder of RD4AD with Mamba-based components.

\paragraph{Multi-Class Setting.}
The ``one-category-one-model'' assumption of traditional VAD is unrealistic in production lines, where dozens of products share a single inspection line. The multi-class setting is more realistic: a single model must separate normal from anomalous samples across categories without knowing the category of the test image.
Many recent methods target this setting, including TS methods such as Dinomaly \cite{dinomaly} and MambaAD \cite{he2024mambaad}.
Although not designed for it, STFPM, RD4AD, and DeSTSeg can easily operate in the multi-class setting by training on the union of the categories.
Among the non-TS methods, DiAD \cite{he2024diffusion} uses diffusion-based reconstruction, which is expensive since it operates in image space.

\paragraph{Continual Learning.}
Continual learning (CL) adds that categories arrive sequentially and that previous data are no longer accessible. CL approaches fall into three main families: replay, regularization, and parameter isolation.
\\
\textbf{Replay} mitigates catastrophic forgetting by reusing information from past tasks, either by interleaving new data with a small replay buffer of past samples or by storing their features (feature memory). 
Replaying past images is the simplest strategy and the first one applied to VAD \cite{bugarin2024unveiling}.
It is compatible with most methods, but it requires additional memory and the ability to store past data, which scales poorly in the long term. 
Feature-memory approaches, namely PatchCoreCL \cite{bugarin2024unveiling}, UCAD \cite{liu2024unsupervised}, and ContCore \cite{jung2026memory}, are memory-bank methods built on pretrained features: they store features rather than images and incrementally expand the memory bank as new tasks arrive.
\\
\textbf{Regularization} methods such as EWC~\cite{kirkpatrick2017overcoming} and LwF~\cite{li2017learning} constrain how the model changes while learning a new task, and have been applied to VAD: IUF \cite{tang2024incremental} uses a regularization of the parameter updates, weighted by the importance of each channel for previous objects, while CDAD \cite{li2025one} projects the gradients onto the subspace orthogonal to the features of previous tasks to preserve previous knowledge in a diffusion-based detector.
\\
\textbf{Parameter isolation} \cite{aljundi2017expert} allocates dedicated parameters to each task, which prevents forgetting by construction and requires no past data, but the memory can grow quickly if the method is not designed carefully. 
Parameter-efficient fine-tuning (PEFT) partially addresses this: prompt-based methods \cite{wang2022sprompts, wang2022dualprompt, wang2022learning} and LoRA-based methods such as SD-LoRA~\cite{wu2025sd} train small task-specific modules on a frozen model and select them at inference.
In CL VAD, UCAD \cite{liu2024unsupervised} follows this strategy, combining prompts with key-query matching and a memory of features. 
AdapTS belongs to this family but, unlike the methods above, it relies on lightweight residual adapters and stores no feature memory, only one prototype per task. Table~\ref{tab:cl_vad} positions AdapTS and the continual VAD methods discussed above with respect to their CL family and their VAD approach.
\input{tables/related_cl_vad_small}
\\
However, to the best of our knowledge, no TS method was specifically developed for the continual setting. We therefore propose AdapTS, which addresses both settings within a unified architecture. Like multi-class methods \cite{he2024mambaad, dinomaly}, it serves all categories with a single deployed model, with category-specific adapters selected through task identification. We also target resource-constrained deployment: the goal is a model that is both adaptive and efficient enough to run close to the production line. Few works study CL VAD in resource-constrained settings; \cite{barusco2025memory} applies replay to an edge-oriented STFPM.

%% file: tables/related_cl_vad_small.tex
\begin{table}[tb]
\centering
\scriptsize
\setlength{\tabcolsep}{4pt}
\renewcommand{\arraystretch}{1.05}
\caption{Continual VAD methods compared with AdapTS. None of them stores past images. The classification of the competing methods is ours, adapted from the ContCore paper \cite{jung2026memory}.}
\label{tab:cl_vad}
\resizebox{\columnwidth}{!}{\begin{tabular}{l l l}
\toprule
\textbf{Method} & \textbf{Continual} & \textbf{VAD} \\
\midrule
PatchCoreCL \cite{bugarin2024unveiling} (CVPR'24) & Replay (feature memory) & Memory bank \\
UCAD \cite{liu2024unsupervised} (AAAI'24) & \begin{tabular}[t]{@{}l@{}}Param. isolation (prompts)\\+ replay (feature memory)\end{tabular} & Memory bank \\
IUF \cite{tang2024incremental} (ECCV'24) & Regularization (importance-weighted update) & Reconstruction \\
CDAD \cite{li2025one} (CVPR'25) & Regularization (gradient projection) & Reconstruction \\
ContCore \cite{jung2026memory} (BMVC'26) & Replay (feature memory) & Memory bank \\
\midrule
\textbf{AdapTS (ours)} & \textbf{Param. isolation (adapters)} & \textbf{Teacher-student} \\
\bottomrule
\end{tabular}}
\end{table}

%% file: sec/methodology.tex
\begin{figure*}[!thbp]
    \centering
    \includegraphics[width=\textwidth]{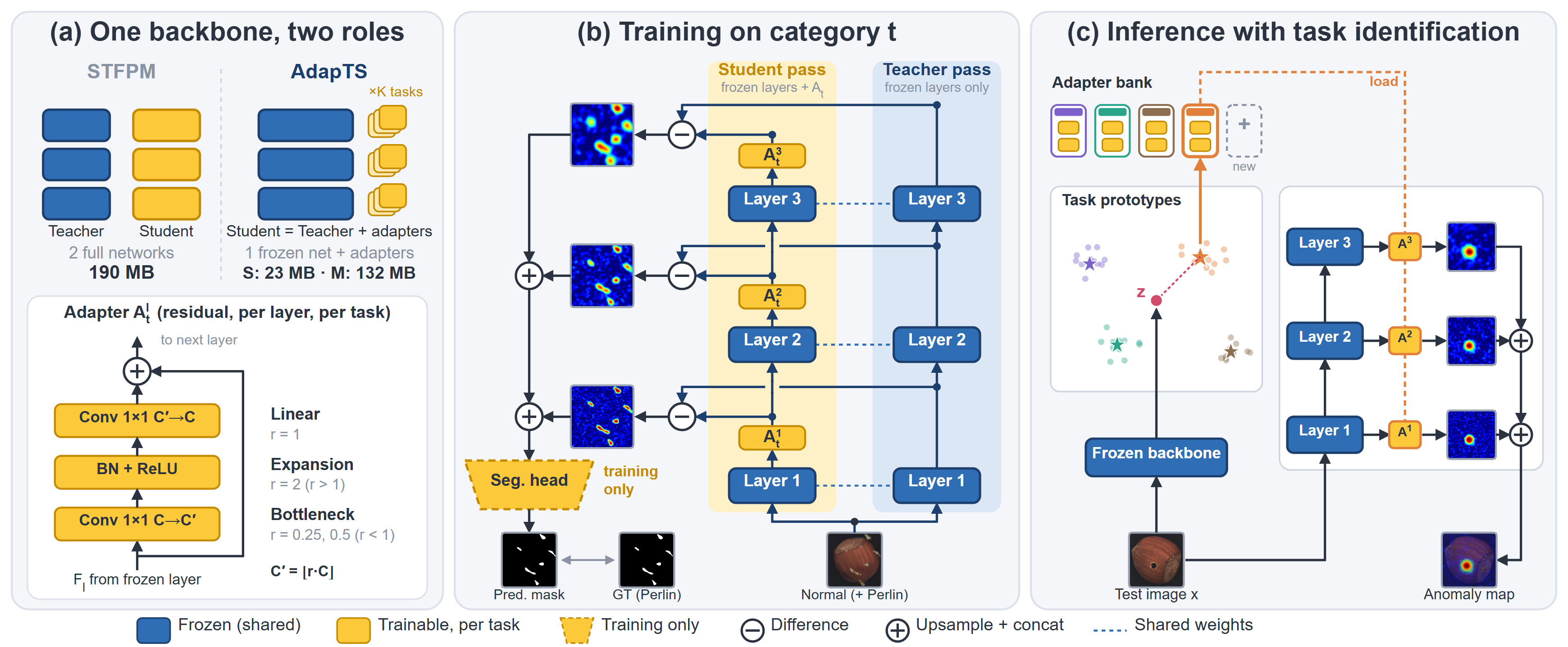}
    \caption{Overview of \textbf{AdapTS}. Blue blocks are frozen and shared, yellow blocks are trainable and task-specific. (a)~A single frozen backbone acts as teacher and, with residual adapters inserted after its layers, as student. Adapters are drawn at three layers for clarity: AdapTS-M uses layers 2 and 3 and AdapTS-S layer 2 (AdapTS-M: 132~MB, AdapTS-S: 23~MB). (b)~During training on category $t$, student--teacher differences are fed to a segmentation head supervised with synthetic Perlin anomalies; the head is used only for training. (c)~At inference, the descriptor of the test image is matched to the task prototypes and the adapters of the closest task are loaded; the anomaly map comes from the student--teacher discrepancy alone.}
        \label{fig:method}
\end{figure*}

\section{Method}
\label{sec:method}

AdapTS builds on STFPM, summarized in Sec.~\ref{subsec:STFPM}. We then describe the adapters that replace the separate student (Sec.~\ref{subsec:adapters}), the segmentation-guided training (Sec.~\ref{subsec:segmentation}), and the task identification mechanism used in the multi-class and continual settings (Sec.~\ref{subsec:task_identification}).

\subsection{Student-Teacher Feature Pyramid Matching for Anomaly Detection}
\label{subsec:STFPM}

STFPM~\cite{wang2021studentteacherfeaturepyramidmatching} uses a student network $S$ and a teacher network $T$ with the same architecture. The student, randomly initialized, is trained to replicate the multi-scale features of the frozen pretrained teacher on normal images.
Given a training set of normal images $D = \{x_1,..., x_n\}$, with $x_i \in \mathbb{R}^{C \times H \times W}$, we extract the feature maps $F_T^l(x), F_S^l(x) \in \mathbb{R}^{C_l \times H_l \times W_l}$ at a set of layers $L$, where $H_l$ and $W_l$ denote the spatial dimensions at layer $l$, and the student minimizes
\begin{equation}
\label{eq:stfpm_loss}
\mathcal{L}_{\text{STFPM}} = 
\sum_{l \in L} 
\frac{1}{H_l W_l} 
\left\| \phi(F_T^l(x)) - \phi(F_S^l(x)) \right\|_2^2,
\end{equation}
where $\phi(\cdot)$ is channel-wise normalization. At inference, the anomaly map is obtained from the differences between student and teacher feature maps, upsampled to the input resolution.

\subsection{Proposed Approach}
\label{subsec:adapters}

STFPM requires a full, randomly initialized copy of the teacher, which doubles the model size. AdapTS instead augments the teacher itself with lightweight adapters inserted after its feature-extraction layers, turning the augmented teacher into its own student (Fig.~\ref{fig:method}). The adapters process the teacher feature maps and pass them through the rest of the network, creating a discrepancy with respect to the plain teacher pass that is minimized on normal images with $\mathcal{L}_{\text{STFPM}}$. Only the adapters are trained, so the additional parameters are a small fraction of those of an STFPM student. Besides acting as the student, the adapters store the knowledge acquired on each category: they play the role of a replay buffer, as the memory of previous tasks, but without storing any past data.

\noindent To formalize the behavior of the developed adapters, we first define the following common notations: $\mathbf{x} \in \mathbb{R}^{C \times H \times W}$ is the input tensor, where $C$ is the number of channels and $H, W$ are the spatial dimensions; the symbol $*$ denotes the spatial convolution operator with a $1 \times 1$ kernel; $\mathbf{W}_1 \in \mathbb{R}^{C_{\text{latent}} \times C \times 1 \times 1}$, $\mathbf{b}_1$ and $\mathbf{W}_2 \in \mathbb{R}^{C \times C_{\text{latent}} \times 1 \times 1}$, $\mathbf{b}_2$ represent the weights and biases of the first and second convolutions, respectively; $\text{BN}(\cdot)$ represents the spatial Batch Normalization operation; and $\sigma(\cdot)$ is the ReLU activation function, defined as $\sigma(z) = \max(0, z)$.
All adapters share a unified residual architecture governed by a single equation, 
where the input $\mathbf{x}$ is added to the output of two sequential $1\times1$ 
convolutions with an intermediate Batch Normalization and non-linearity:
\begin{equation}
\label{eq:adapter}
    \mathbf{y} = \mathbf{x} + \mathbf{W}_2 * \sigma\Big(\text{BN}\big(\mathbf{W}_1 * \mathbf{x} + \mathbf{b}_1\big)\Big) + \mathbf{b}_2
\end{equation}
The three adapter variants differ solely in the choice of the latent dimensionality $C_{\text{latent}} = \max\!\left(1,\lfloor C \cdot r \rfloor\right)$, controlled by a scaling factor $r$.
\textbf{Linear adapter}: $r = 1$. The channel dimension is preserved throughout, so both projection matrices are square. \textbf{Expansion adapter}: $r > 1$. Features are projected into a higher-dimensional space to improve intermediate separability between student and teacher representations, before being projected back. We use $r = 2$. \textbf{Bottleneck adapter}: $0 < r < 1$. Features are compressed into a lower-dimensional bottleneck to encourage the adapter to focus on reproducing only the most informative teacher features, suppressing spurious noise. We analyze two bottleneck variants, BN25 and BN50, with $r = 0.25$ and $r = 0.5$, respectively.
To further reduce the memory footprint of AdapTS, the adapters can be INT-8 quantized using a post-training quantization procedure without impacting the performance, as shown in Table~\ref{tab:results-multi-eff} and discussed in Sec.~\ref{sec:ablation}.
In the Multi-Class and Continual scenarios, a separate set of adapters is trained for each category and selected at inference time through the task identification described in Sec.~\ref{subsec:task_identification}.

\subsection{Segmentation guided training}
\label{subsec:segmentation}

Following~\cite{zhang2023destseg}, we guide the adapter training with an auxiliary segmentation task. Since the training set contains only normal images, we generate synthetic anomalies and their masks with Perlin noise and random textures. The student-teacher difference maps at the selected layers are upsampled to a common resolution, concatenated channel-wise, and fed to a segmentation head that predicts a pixel-wise anomaly probability map. We consider three heads of increasing capacity (Simple, Medium, and Complex; details in Sec. \ref{subsec:settings_models}).
\\
Unlike~\cite{zhang2023destseg}, where the segmentation network produces the final anomaly map at inference, our head is used only during training and discarded afterwards: at inference, the anomaly map is computed directly from the student-teacher discrepancy, with no additional cost. In AdapTS, the head only provides a supervision signal that shapes the adapters: since the segmentation loss is back-propagated through the difference maps, the adapters learn to produce low discrepancies on normal regions and high, spatially coherent discrepancies on anomalous ones.

The ground-truth anomaly mask is downsampled to match the output size $(H_{s}, W_{s})$. Denoting the predicted probability map as $\hat{Y}$ and the downsampled mask as $M$, the segmentation loss is the sum of a focal term and an L1 term:
\begin{equation}
    \mathcal{L}_{\text{seg}} = \frac{1}{H_s W_s} \sum_{i,j=1}^{H_s, W_s} \Big[ - (1 - p_{ij})^\gamma \log(p_{ij}) + |M_{ij} - \hat{Y}_{ij}| \Big],
\end{equation}
where $p_{ij} = M_{ij}\hat{Y}_{ij} + (1 - M_{ij})(1 - \hat{Y}_{ij})$ and $\gamma$ is the focusing parameter ($\gamma = 4$). The focal term helps the model to focus on the minority category and difficult samples, while the L1 term is employed to improve the sparsity of the output so that the segmentation mask’s boundaries are more distinct. 
The final loss for training the adapters is thus:
\begin{equation}
\label{eq:adapt_loss}
    \mathcal{L} = \mathcal{L}_{\text{STFPM}} + \lambda \mathcal{L}_{\text{seg}}
\end{equation}
$\lambda = 100$ is set in the experiments in order to match the two loss components values.
Figure~\ref{fig:method} visually summarizes the key components of AdapTS.

\davide{\textbf{Avoiding the identity solution.} Since the adapter in Eq.~\eqref{eq:adapter} is residual, $\mathcal{L}_{\text{STFPM}}$ has a trivial global minimum: with $\mathbf{W}_2 = 0$ and $\mathbf{b}_2 = 0$ the student coincides with the teacher, the discrepancy vanishes on anomalous images as well, and detection performance drops.
Encoder-decoder TS methods face the same shortcut~\cite{you2022unified} and counter it with a compact~\cite{deng2022anomaly} or noisy~\cite{dinomaly} bottleneck. AdapTS has no such constraint, but two factors keep it away from the identity. First, adapters are commonly initialized near the identity~\cite{houlsby2019parameter}; in AdapTS, a zero $\mathbf{W}_2$ places them exactly there, where $\mathcal{L}_{\text{STFPM}}$ has zero gradient and training stalls. Any non-zero initialization we tested (orthogonal and Xavier), even one scaled close to zero, avoids this, and we use the orthogonal one. Once away from the identity, training does not return to it: the adapters reduce the discrepancy on normal images while remaining different from the teacher, so anomalies still produce a high discrepancy (Sec.~\ref{subsec:results_ablation}). Second, the segmentation loss makes the identity suboptimal for the full objective: with zero discrepancy, the head cannot localize the synthetic anomalies.}

\subsection{Task identification}
\label{subsec:task_identification}

A key advantage of our approach is the ability to train task-specific adapters tailored to different input data distributions. Let $\mathcal{T} = \{1, \dots, K\}$ denote the set of tasks (\ie, the object categories in the dataset). For each task $t \in \mathcal{T}$, we define a prototype $\mathbf{p}_t \in \mathbb{R}^{D}$ computed from the features of the frozen teacher at the deepest layer used by the model, $l_{\max} = \max(L)$ (\eg, layer 2 for AdapTS-S and layer 3 for AdapTS-L/M). Given the feature map $F_T^{l_{\max}}(x) \in \mathbb{R}^{W_{l_{\max}} \times H_{l_{\max}} \times C_{l_{\max}}}$, we obtain a global descriptor via global average pooling (GAP), $\mathbf{z}(x) = \mathrm{GAP}\big(F_T^{l_{\max}}(x)\big) \in \mathbb{R}^{D}$, with $D = C_{l_{\max}}$.

Given the training set $\mathcal{D}_t = \{x_i^{(t)}\}_{i=1}^{N_t}$ of task $t$, the prototype is the mean descriptor
\[
\mathbf{p}_t = \frac{1}{N_t} \sum_{i=1}^{N_t} \mathbf{z}\big(x_i^{(t)}\big).
\]
At inference time, a test image $x$ is assigned to the task with the closest prototype in Euclidean distance,
\[
\hat{k} = \arg\min_{t \in \mathcal{T}} \big\lVert \mathbf{z}(x) - \mathbf{p}_t \big\rVert_2,
\]
and the adapters of task $\hat{k}$ are loaded. 
Since $\mathbf{z}(x)$ reuses the teacher features needed for the discrepancy, task identification adds no forward pass: the teacher is run once, and a second pass through the adapted network of task $\hat{k}$ provides the student features. Only one prototype per task ($K$ in total) needs to be stored.

%% file: sec/settings.tex
\section{Experimental Setting}
\label{sec:experimental_setting}

\subsection{Implementation Details}
\label{subsec:implementation_details}

For a fair comparison, all the VAD models use the WideResNet-50 pretrained on ImageNet as teacher, except Dinomaly, MambaAD and DiAD, since a WideResNet-50 backbone does not fit their design. Adapters and segmentation head are trained jointly for 30 epochs with SGD (batch size 32, learning rate 0.1 with cosine annealing to $10^{-3}$). The task prototypes (Sec.~\ref{subsec:task_identification}) use the teacher features at layer 3 (AdapTS-L/M) or layer 2 (AdapTS-S). As in Dinomaly, the image-level score of AdapTS is the mean of the top 1\% values of the anomaly map, while the ablations use the maximum (Sec.~\ref{sec:ablation}). The main results are averaged over three seeds. We train the other methods using their official code and default hyperparameters (see the Supplementary Material).

\subsection{Datasets}
MVTec AD \cite{bergmann2019mvtec} contains high-resolution images of 15 categories (10 objects and 5 textures), with a defect-free training set and a test set with normal and anomalous samples and pixel-level annotations. VisA \cite{zou2022spotthedifferenceselfsupervisedpretraininganomaly} comprises 10,821 high-resolution color images of 12 objects in 3 domains.

\subsection{Models}
\label{subsec:settings_models}
We evaluate three variants of AdapTS: \textbf{AdapTS-L}, with adapters at layers 2 and 3; \textbf{AdapTS-M}, with INT-8 quantized adapters at layers 2 and 3; and \textbf{AdapTS-S}, the lightest, with only the INT-8 quantized adapter at layer 2. We compare them with two groups of methods, as in Fig.~\ref{fig:tradeoff}. \emph{Adapted} methods are not designed for continual learning: STFPM~\cite{wang2021studentteacherfeaturepyramidmatching}, RD4AD~\cite{deng2022anomaly}, DeSTSeg~\cite{zhang2023destseg}, Dinomaly~\cite{dinomaly} and MambaAD~\cite{he2024mambaad} are evaluated in both scenarios (with a replay buffer in the Continual one), while DiAD~\cite{he2024diffusion} only in the Multi-Class one. \emph{Designed for CL} methods are UCAD~\cite{liu2024unsupervised}, IUF~\cite{tang2024incremental}, CDAD~\cite{li2025one}, ContCore~\cite{jung2026memory} and PatchCoreCL~\cite{bugarin2024unveiling} (Table~\ref{tab:cl_vad}).

We evaluate three segmentation heads of increasing capacity (Simple: a $1\times1$ convolution; Medium: a $1\times1$--$3\times3$--$1\times1$ stack with 64 hidden channels; Complex: two 256-channel residual blocks followed by an ASPP module, as in DeSTSeg~\cite{zhang2023destseg}) and adopt the Complex one, which performs best (Sec.~\ref{sec:ablation}).

\subsection{Scenarios and Evaluation Metrics}
\label{subsec:evaluation_metrics}

\textbf{Scenarios.} All models are evaluated in two scenarios: \textit{Multi-Class}, where a single model is deployed for all categories, and \textit{Continual}, where categories arrive sequentially as a stream. This is the standard protocol of continual VAD~\cite{liu2024unsupervised, tang2024incremental, li2025one, jung2026memory}: categories arrive one at a time, data from previous categories is not available, and the category of a test image is not given at inference. In the Continual scenario, the adapted methods use a replay buffer of 100 randomly selected samples. Metrics are computed per category after training on all categories and then averaged. 
\\
\textbf{Metrics.} Detection is evaluated with the image-level AUROC (I-ROC), and localization with the pixel-level AUROC (P-ROC), F1 score (P-F1), and AP. We include P-F1 and AP because P-ROC discriminates little between methods: in Table~\ref{tab:results-multi-eff} it varies by only 4 points, while AP varies by 15. Efficiency is measured by the total memory (pretrained teacher plus any additional memory: student or decoder, replay buffer, adapters and task prototypes) and by the VRAM and FLOPs at training and inference. In the Continual scenario, we also report the Average Forgetting F on P-ROC (defined in the Supplementary Material). The results of UCAD, IUF, CDAD and ContCore are taken from~\cite{jung2026memory}, whose main tables report I-ROC, P-ROC and the forgetting on P-ROC; P-F1 and pixel-level AP are thus not available for them.

\input{tables/table_results_multiclass_eff}
\input{tables/table_results_continual}

%% file: tables/table_results_multiclass_eff.tex
\begin{table}[tb]
\centering
\scriptsize
\setlength{\tabcolsep}{3pt}
\renewcommand{\arraystretch}{1.05}
\caption{Comparison of anomaly detection models on \textbf{MVTec AD} (15 categories) in the \textbf{Multi-Class scenario}. \textbf{Bold} indicates the best result. Tot: total memory; Inf: inference; Tr: training.}
\label{tab:results-multi-eff}
\resizebox{\columnwidth}{!}{
\begin{tabular}{l c >{\columncolor{gray!15}}c c >{\columncolor{gray!15}}c@{\hspace{8pt}} c@{\hspace{8pt}} >{\columncolor{gray!15}}c >{\columncolor{gray!15}}c c c}
\toprule
& \multicolumn{1}{c}{\textbf{Image}} & \multicolumn{3}{c@{\hspace{8pt}}}{\textbf{Pixel}} & \multicolumn{1}{c@{\hspace{8pt}}}{\textbf{Mem (MB)}} & \multicolumn{2}{c}{\textbf{FLOPs (G)}} & \multicolumn{2}{c}{\textbf{VRAM (MB)}} \\
\cmidrule(lr){2-2} \cmidrule(lr){3-5} \cmidrule(lr){6-6} \cmidrule(lr){7-8} \cmidrule(lr){9-10}
\rowcolor{white} \textbf{Model} & \textbf{I-ROC} & \textbf{P-ROC} & \textbf{P-F1} & \textbf{AP} & \textbf{Tot} & \textbf{Inf} & \textbf{Tr} & \textbf{Inf} & \textbf{Tr} \\
\midrule
AdapTS-L & 0.89 & 0.95 & 0.52 & 0.50 & 245 & 44 & 90 & 169 & 358 \\
AdapTS-M & 0.89 & 0.95 & 0.52 & 0.50 & 132 & 44 & 90 & 161 & 351 \\
AdapTS-S & 0.87 & 0.92 & 0.45 & 0.43 & \textbf{23} & \textbf{35} & \textbf{69} & \textbf{80} & \textbf{172} \\
\midrule
STFPM    & 0.78 & 0.95 & 0.45 & 0.44 & 190 & 37 & 74 & 335 & 604 \\
RD4AD    & 0.85 & 0.93 & 0.48 & 0.41 & 615 & 48 & 147 & 970 & 1242 \\
DeSTSeg  & 0.86 & 0.92 & \textbf{0.54} & \textbf{0.55} & 1215 & 187 & 977 & 2646 & 3241 \\
Dinomaly & 0.93 & 0.93 & 0.49 & 0.40 & 561 & 66 & 132 & 592 & 823 \\
MambaAD  & \textbf{0.95} & \textbf{0.96} & 0.52 & 0.48 & 330 & 38 & 99  & 750 & 1540 \\
DiAD     & \textbf{0.95} & \textbf{0.96} & 0.53 & 0.49 & 5430 & 3375 & 896 & 5580 & 20050 \\
\bottomrule
\end{tabular}
}
\end{table}

%% file: tables/table_results_continual.tex
\begin{table}[tb]
\centering
\scriptsize
\setlength{\tabcolsep}{4pt}
\renewcommand{\arraystretch}{1.05}
\caption{Comparison on \textbf{MVTec AD} in the \textbf{Continual scenario}. \textbf{Bold}: best. F: average forgetting on P-ROC. Tot: total memory, including the replay buffer (100 samples, 14~MB) for the adapted methods. *Results reported in~\cite{jung2026memory}.}
\label{tab:results-cont}
\begin{tabular}{l c >{\columncolor{gray!15}}c c >{\columncolor{gray!15}}c c c}
\toprule
& \multicolumn{1}{c}{\textbf{Image}} & \multicolumn{3}{c}{\textbf{Pixel}} & \multicolumn{1}{c}{\textbf{F}} & \multicolumn{1}{c}{\textbf{Mem (MB)}} \\
\cmidrule(lr){2-2} \cmidrule(lr){3-5} \cmidrule(lr){6-6} \cmidrule(lr){7-7}
\rowcolor{white} \textbf{Model} & \textbf{I-ROC} & \textbf{P-ROC} & \textbf{P-F1} & \textbf{AP} & \textbf{\%} & \textbf{Tot} \\
\midrule
AdapTS-L   & 0.89 & 0.95 & \textbf{0.52} & \textbf{0.50} & \textbf{0.1} & 245 \\
AdapTS-M   & 0.89 & 0.95 & \textbf{0.52} & \textbf{0.50} & \textbf{0.1} & 132 \\
AdapTS-S   & 0.87 & 0.92 & 0.45 & 0.43 & \textbf{0.1} & \textbf{23} \\
\midrule
STFPM      & 0.78 & 0.91 & 0.41 & 0.36 & 12 & 204 \\
RD4AD      & 0.94 & 0.95 & 0.51 & 0.47 & 9 & 629 \\
DeSTSeg    & 0.80 & 0.92 & 0.46 & 0.45 & 10 & 1229 \\
Dinomaly   & 0.93 & 0.95 & 0.48 & 0.41 & 7 & 575 \\
MambaAD    & 0.90 & 0.95 & 0.48 & 0.43 & 8 & 344 \\
\midrule
UCAD*      & 0.93 & 0.83 & -- & -- & 0.6 & 276 \\
IUF*       & 0.72 & 0.79 & -- & -- & 6.6 & 144 \\
CDAD*      & 0.76 & 0.90 & -- & -- & 5.7 & 4100 \\
ContCore*  & \textbf{0.99} & \textbf{0.97} & -- & -- & 0.2 & 177 \\
PatchCoreCL (30k) & 0.98 & 0.93 & 0.48 & 0.45 & 1.7 & 300 \\
\bottomrule
\end{tabular}
\end{table}

%% file: sec/results.tex
\section{Results}
\label{sec:results}

We compare AdapTS with state-of-the-art methods on MVTec AD in the Multi-Class and Continual scenarios (Sec.~\ref{subsec:results_multi_continual_scenarios}), evaluate a MobileNetV2 backbone for resource-constrained deployment (Sec.~\ref{subsec:results_mobilenet}), extend the comparison to VisA (Sec.~\ref{subsec:results_visa}) and report the ablation studies (Sec.~\ref{subsec:results_ablation}).

\subsection{Performance Comparison across Scenarios}
\label{subsec:results_multi_continual_scenarios}

\begin{figure}[!thbp]
    \centering
    \includegraphics[width=\columnwidth]{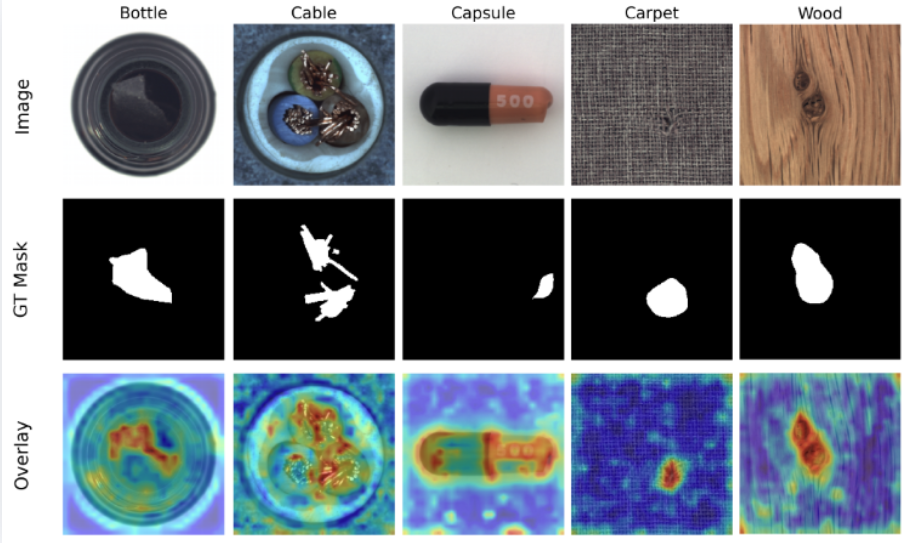}
    \caption{AdapTS anomaly detection examples on the MVTec Dataset, multi-class setting. The first row contains an anomalous image per category, the second row the ground-truth anomaly mask and the third row the anomaly heatmap produced by AdapTS.}
    \label{fig:qualitative}
\end{figure}

\textbf{Multi-Class scenario}. Table~\ref{tab:results-multi-eff} and Fig.~\ref{fig:tradeoff} compare AdapTS with the adapted methods, here standard multi-class models. AdapTS-S is the most efficient method, and AdapTS-M is competitive in detection with less memory than all the others. Compared with STFPM, AdapTS-M improves I-ROC by 11 points and AP by 6 while using less memory, and it outperforms RD4AD on all metrics with almost five times less memory. Dinomaly, MambaAD and DiAD reach a higher I-ROC, between 0.93 and 0.95, but a lower AP than AdapTS-M, and they need between 2.5 and 41 times its memory. Only DeSTSeg has a higher AP (0.55), but it needs about nine times the memory. With only 23~MB, AdapTS-S still has a higher I-ROC than STFPM, RD4AD and DeSTSeg, with the lowest FLOPs and VRAM of all methods. Qualitative examples are shown in Fig.~\ref{fig:qualitative}.

\textbf{Continual scenario}. Table~\ref{tab:results-cont} compares AdapTS with the adapted methods (first block) and those designed for CL (second block). By design, AdapTS obtains the same results as in the Multi-Class scenario: each new category adds a set of adapters and a prototype, and previous adapters are never updated, so forgetting arises only from task identification errors (0.1\%, consistent with an accuracy above 99\%, see the Supplementary Material). It also stores no past data, whereas the adapted methods need a replay buffer.
AdapTS-M and AdapTS-L reach the best P-F1 and AP (0.50 vs. 0.47 for the best competitor, RD4AD) among all the adapted methods, which forget between 7\% and 12\%.
RD4AD, Dinomaly and MambaAD obtain a higher I-ROC, between 0.90 and 0.94, but need between 2.6 and 4.8 times the memory of AdapTS-M.

Methods designed for CL are strong at the image level: ContCore, PatchCoreCL and UCAD reach an I-ROC between 0.93 and 0.99, against 0.89 for AdapTS-M. ContCore also has the highest P-ROC (0.97) and a very low forgetting (0.2\%), with 177~MB of memory, but it does not report P-F1 or pixel-level AP.
PatchCoreCL is the only one for which we can compare localization, and AdapTS-M is better in P-F1 and AP, with less forgetting (0.1\% vs. 1.7\%) and 2.3 times less memory. IUF and CDAD have a lower I-ROC and a higher forgetting, about 6\%, and CDAD requires 4.1~GB. In summary, AdapTS is not the best at image-level detection, but it has the best P-F1 and AP among the continual methods that report them, a very low forgetting, and 23--132~MB of memory, below all the other methods (the lightest, IUF, needs 144~MB).

\davide{\textbf{Memory growth.} Each new category adds 2.5~MB to AdapTS-M and 0.5~MB to AdapTS-S, the memory of about 18 and 4 replay images, respectively. If a replay buffer grows at the rate of our experiments (about 7 images, or 0.9~MB, per category), AdapTS-M needs less memory than STFPM up to about 60 categories and than Dinomaly up to about 300, while AdapTS-S needs less memory than all the other methods for any number of categories (see the Supplementary Material).}

\subsection{MobileNet Backbone}
\label{sec:mobilenet_backbone}
\label{subsec:results_mobilenet}

\input{tables/results_mobilenet}

To demonstrate its suitability for resource-constrained deployment, we evaluate a lightweight AdapTS-L built on a MobileNetV2 backbone pretrained on ImageNet, against the edge-oriented variant of STFPM of~\cite{barusco2025memory}, which also uses a MobileNet. Table~\ref{tab:results-mobilenet} reports the Continual scenario. On both datasets, AdapTS outperforms STFPM on every metric with a fraction of its memory: 7 and 6~MB on MVTec and VisA, against 18~MB. On MVTec the gain is modest at the image level but larger at the pixel level, with AP rising from 0.38 to 0.48. On the harder VisA the gain widens, with 13 more points of I-ROC, 15 of P-F1 and 14 of AP. AdapTS thus remains effective in highly resource-constrained scenarios, and its advantage grows on the harder dataset.

\input{tables/results_visa}

\subsection{VisA Dataset}
\label{sec:adapters_visa}
\label{subsec:results_visa}

Table~\ref{tab:results-visa} shows the Multi-Class scenario on VisA. AdapTS-M outperforms STFPM on I-ROC and AP with less memory. Compared with RD4AD, it has a higher AP (0.29 vs. 0.26) but a slightly lower I-ROC (0.83 vs. 0.85), with about five times less memory. Dinomaly and MambaAD are the most accurate TS methods, with an I-ROC of 0.94 and 0.93, but they need between 2.6 and 4.5 times the memory of AdapTS-M. With only 22~MB, AdapTS-S has a higher I-ROC than STFPM and is close to RD4AD (0.84 vs. 0.85) with 28 times less memory. AdapTS thus offers competitive multi-class detection at a fraction of the memory of the most accurate methods, a practical choice for resource-constrained deployment.

\input{tables/table_layers_ablation}
\subsection{Ablation Studies}
\label{sec:ablation}
\label{subsec:results_ablation}

The ablation studies are conducted on MVTec AD in the single-model setting, where a separate model is trained for each category. Table~\ref{tab:adapters_ablation} is organized in three blocks: adapter architecture, segmentation head, and layer placement.
\input{tables/table_adapters_ablation}

\noindent\textbf{Adapter architecture.}
As shown in the first block of Table~\ref{tab:adapters_ablation}, compressing the features harms the representation: compared to the Linear adapter, BN25 and BN50 lose up to 6 points of I-ROC and 17 points of AP. Conversely, the Expansion adapter improves I-ROC by only 2 points, with no gain in AP, while doubling the memory. Therefore, the Linear adapter achieves the best trade-off.

\noindent\textbf{Segmentation head.}
In the second block, segmentation-guided training improves the Linear adapter, and the gain grows with the head capacity: the Complex head adds 4 points of I-ROC and 5 points of AP. Since the head is discarded after training, these gains come at no cost in memory or inference. Using the head also at inference, instead, lowers I-ROC from 0.87 to 0.83 and AP from 0.48 to 0.45, confirming its role as a training signal for the adapters rather than as an anomaly detector.

\noindent\textbf{Layer placement.} The third block applies Linear Seg Complex to different layer combinations. Removing layer 1 slightly improves the results, so layers 2 and 3 form AdapTS-L; layer 3 matters most, since removing it costs 8 points of I-ROC and 12 of AP. A single adapter at layer 2 stays competitive (I-ROC 0.85) with five times less memory, and is used in AdapTS-S. The last two rows show that the top-1\% score used in all main results improves I-ROC by 1 point for layers 2, 3 and by 2 points for layer 2 over the maximum.

\noindent\textbf{Quantization.} Quantizing the adapters to INT-8, which turns AdapTS-L into AdapTS-M, leaves all the results unchanged (Table~\ref{tab:results-multi-eff}), while cutting the memory of the adapters of one category by four times (from 10.03 to 2.51~MB, Table~\ref{tab:adapters_ablation}) and the total memory from 245 to 132~MB.

\noindent\textbf{Task identification.}
The prototype-based technique selects the correct adapters for more than 99\% of the test images on both MVTec and VisA,  making task identification errors a negligible source of forgetting (more in the Supplementary Material).

\noindent\davide{\textbf{Identity collapse.} The adapters do not converge to the identity of Sec.~\ref{subsec:segmentation}. First, training leaves the identity even when it starts close to it: with $\mathbf{W}_2$ orthogonally initialized and scaled by $1/100$ and no segmentation head, $\lVert\mathbf{W}_2\rVert$ grows by 4--8$\times$ with a learning rate of 0.01 (ten times lower than ours) and by 20--50$\times$ with 0.1, in all the runs with 2 learning rates. Second, collapse does not occur even without the head: the Linear adapter reaches 0.83 I-ROC (Table~\ref{tab:adapters_ablation}), far from the 0.5 of the identity. The head is thus not required to avoid the identity, but it improves detection and localization (4 points of I-ROC, 5 of AP) and removes the identity from the minimizers of the objective.}

%% file: tables/results_mobilenet.tex
\begin{table}[tb]
\centering
\scriptsize
\setlength{\tabcolsep}{3pt}
\renewcommand{\arraystretch}{1.05}
\caption{Comparison of STFPM and AdapTS-L in the Continual scenario with MobileNetV2 as backbone. \textbf{Bold} values indicate the best result per metric within each dataset.}
\label{tab:results-mobilenet}
\begin{tabular}{l l
    >{\columncolor{white}}c
    >{\columncolor{lightgray}}c
    >{\columncolor{white}}c
    >{\columncolor{lightgray}}c
    >{\columncolor{white}}c}
\toprule
\textbf{Dataset} & \textbf{Model}
& \multicolumn{1}{c}{\textbf{Image-level}}
& \multicolumn{3}{c}{\textbf{Pixel-level}}
& \multicolumn{1}{c}{\textbf{Memory (MB)}} \\
\cmidrule(lr){3-3}
\cmidrule(lr){4-6}
\cmidrule(lr){7-7}
\rowcolor{white} & 
& \textbf{I-ROC}
& \textbf{P-ROC} 
& \textbf{P-F1} 
& \textbf{AP} 
& \textbf{Total} \\
\midrule
\multirow{2}{*}{\parbox{1.2cm}{ \centering MVTec\\{\normalfont\tiny(15 classes)}}} 
& STFPM  & 0.86          & 0.91          & 0.42          & 0.38          & 18 \\
& AdapTS-L & \textbf{0.89} & \textbf{0.93} & \textbf{0.49} & \textbf{0.48} & \textbf{7} \\
\midrule
\multirow{2}{*}{\parbox{1.2cm}{\centering VisA\\{\normalfont\tiny(12 classes)}}}  
& STFPM  & 0.74          & 0.92          & 0.20          & 0.15          & 18 \\
& AdapTS-L & \textbf{0.87} & \textbf{0.95} & \textbf{0.35} & \textbf{0.29} & \textbf{6} \\
\bottomrule
\end{tabular}
\end{table}

%% file: tables/results_visa.tex
\begin{table}[tb]
\centering
\scriptsize
\setlength{\tabcolsep}{3pt}
\renewcommand{\arraystretch}{1.05}
\caption{Comparison on \textbf{VisA} (12 categories) in the \textbf{Multi-Class scenario}. \textbf{Bold}: best. *Results reported in~\cite{jung2026memory}.}
\label{tab:results-visa}
\begin{tabular}{l
    >{\columncolor{white}}c
    >{\columncolor{lightgray}}c
    >{\columncolor{white}}c
    >{\columncolor{lightgray}}c
    >{\columncolor{white}}c}
\toprule
& \multicolumn{1}{c}{\textbf{Image-level}}
& \multicolumn{3}{c}{\textbf{Pixel-level}}
& \multicolumn{1}{c}{\textbf{Memory (MB)}} \\
\cmidrule(lr){2-2}
\cmidrule(lr){3-5}
\cmidrule(lr){6-6}
\rowcolor{white} \textbf{Model}
& \textbf{I-ROC}
& \textbf{P-ROC} 
& \textbf{P-F1} 
& \textbf{AP} 
& \textbf{Total} \\
\midrule
AdapTS-L (our)       & 0.83          & 0.96          & 0.35          & 0.29          & 215 \\
AdapTS-M (our)       & 0.83          & 0.96          & 0.34          & 0.29          & 125 \\
AdapTS-S (our)       & 0.84          & 0.94          & 0.33          & 0.26          & \textbf{22} \\
\midrule
STFPM (BMVC 2021)    & 0.79          & 0.95          & 0.36          & 0.23          & 190 \\
RD4AD (CVPR 2022)    & 0.85          & 0.93          & 0.37          & 0.26          & 615 \\
DeSTSeg (CVPR 2023)  & 0.87          & 0.96          & \textbf{0.47} & 0.41          & 1215 \\
Dinomaly (CVPR 2025) & \textbf{0.94} & \textbf{0.99} & 0.45          & \textbf{0.42} & 561 \\
DiAD (AAAI 2024)       & 0.82          & 0.91          & 0.29          & 0.21          & 5430 \\
MambaAD (NeurIPS 2024) & 0.93          & 0.98          & 0.42          & 0.37          & 330 \\
\midrule
UCAD* (AAAI 2024) & 0.78 & 0.77 & -- & -- & 276 \\
IUF* (ECCV 2024) & 0.65 & 0.84 & -- & -- & 144 \\
CDAD* (CVPR 2025) & 0.79 & 0.97 & -- & -- & 4100 \\
ContCore* (BMVC 2026) & \textbf{0.94} & 0.97 & -- & -- & 259 \\
\bottomrule
\end{tabular}
\end{table}

%% file: tables/table_adapters_ablation.tex
\begin{table}[tb]
\centering
\scriptsize
\setlength{\tabcolsep}{3pt}
\renewcommand{\arraystretch}{1.05}
\caption{Ablation on MVTec AD. \textit{Top}: adapter architectures at layers 1, 2, 3. \textit{Middle}: Linear adapter with segmentation heads of increasing capacity; the last row also uses the head at inference. \textit{Bottom}: Linear Seg Complex at different layers (2, 3: AdapTS-L; 2: AdapTS-S). Unless noted, the image-level score is the maximum of the anomaly map; rows marked top-1\% use the mean of its top 1\% values, as in the main results. Memory: adapters of one category, FP32 / INT-8. \textbf{Bold}: best per performance metric within each block.}
\label{tab:adapters_ablation}
\begin{tabular}{l
    >{\columncolor{white}}c
    >{\columncolor{lightgray}}c
    >{\columncolor{white}}c
    >{\columncolor{lightgray}}c
    >{\columncolor{white}}c
    >{\columncolor{lightgray}}c}
\toprule
& \multicolumn{1}{c}{\textbf{Image-level}}
& \multicolumn{3}{c}{\textbf{Pixel-level}}
& \multicolumn{2}{c}{\textbf{Memory [MB]} } \\
\cmidrule(lr){2-2}
\cmidrule(lr){3-5}
\cmidrule(lr){6-7}
\rowcolor{white} \textbf{Model}
& \textbf{I-ROC} 
& \textbf{P-ROC} 
& \textbf{P-F1} 
& \textbf{AP} 
& \textbf{Normal}
& \textbf{Quantized} \\
\midrule
Linear      & 0.83 & 0.90 & 0.44 & \textbf{0.43} & 10.54 & 2.64 \\
BN25        & 0.77 & 0.89 & 0.32 & 0.26 & 2.67  & 0.68 \\
BN50        & 0.78 & 0.90 & 0.35 & 0.30 & 5.30  & 1.33 \\
Expansion   & \textbf{0.85} & \textbf{0.92} & \textbf{0.46} & \textbf{0.43 }& 21.07 & 5.28 \\
\midrule
Linear Seg Simple  & 0.85 & 0.92 & 0.48 & 0.45 & 10.54 & 2.64 \\
Linear Seg Medium  & 0.85 & 0.93 & 0.49 & 0.46 & 10.54 & 2.64 \\
Linear Seg Complex & \textbf{0.87} & \textbf{0.94} & \textbf{0.50} & \textbf{0.48} & 10.54 & 2.64 \\
\begin{tabular}[c]{@{}l@{}}Linear Seg Complex\\(seg head at inference)\end{tabular} & 0.83 & 0.90 & 0.45 & 0.45 & 10.54 & 2.64 \\
\midrule
Layers 1, 2, 3 & 0.87 & 0.94 & 0.50 & 0.48 & 10.54 & 2.64 \\
Layers 1, 2    & 0.79 & 0.91 & 0.40 & 0.36 & 2.51  & 0.63 \\
Layers 2, 3    & 0.88 & \textbf{0.95} & \textbf{0.52} & \textbf{0.50} & 10.03 & 2.51 \\
Layer 2        & 0.85 & 0.92 & 0.45 & 0.43 & \textbf{2.01} & \textbf{0.50} \\
Layers 2, 3 (top-1\%) & \textbf{0.89} & \textbf{0.95} & \textbf{0.52} & \textbf{0.50} & 10.03 & 2.51 \\
Layer 2 (top-1\%)     & 0.87 & 0.92 & 0.45 & 0.43 & \textbf{2.01} & \textbf{0.50} \\
\bottomrule
\end{tabular}
\end{table}

%% file: sec/conclusions.tex
\section{Conclusion}
\label{sec:conclusion}

In this paper, we presented AdapTS, a lightweight teacher-student model for VAD that natively operates in the Multi-Class and Continual scenarios, with a single shared frozen backbone and lightweight trainable adapters. On MVTec AD and VisA, AdapTS is competitive in detection at a fraction of the memory of heavier methods. In the Continual scenario, AdapTS-M achieves the best pixel-level AP and P-F1 with only 0.1\% forgetting and no replay buffer, and with a MobileNetV2 backbone it needs only 6--7~MB. To the best of our knowledge, it is the first TS-based method to unify multi-class and continual learning in a single framework. These results make AdapTS a practical solution for inspection lines with limited resources, where one model must serve many products and new ones are added over time without storing past data.
Like other parameter-isolation methods AdapTS adds parameters for each new category, although at a small cost (0.5--2.5~MB per category for versions S and M). Future work could share adapters across similar categories and add unseen-category detection for online continual learning.